\documentclass[11pt,a4paper]{article}

\usepackage{times}
\usepackage{latexsym}
\usepackage{booktabs}
\usepackage{amsmath,amssymb,amsthm}
\usepackage{tikz}
\usepackage{ulem}
\usetikzlibrary{arrows,shapes,positioning,backgrounds}
\usetikzlibrary{patterns}
\usepackage{templates/acl2017}

\aclfinalcopy 
\def\aclpaperid{134} 

\title{Neural End-to-End Learning for Computational Argumentation Mining}
\author{
Steffen Eger\textsuperscript{\textdagger\textdaggerdbl}, Johannes Daxenberger\textsuperscript{\textdagger}, Iryna Gurevych\textsuperscript{\textdagger\textdaggerdbl}\\
\textsuperscript{\textdagger}Ubiquitous Knowledge Processing Lab (UKP-TUDA)\\
Department of Computer Science, Technische Universität Darmstadt\\
\textsuperscript{\textdaggerdbl}Ubiquitous Knowledge Processing Lab (UKP-DIPF)\\
German Institute for Educational Research and Educational Information\\
\url{http://www.ukp.tu-darmstadt.de}
}

\date{}

\begin{document}
\maketitle
\begin{abstract}
  We investigate neural techniques for end-to-end computational argumentation mining (AM). 
  We frame AM both as a token-based dependency parsing and as a token-based sequence tagging problem, including a multi-task learning setup. 
  Contrary to models that operate on the argument component level, we find that framing AM as dependency parsing 
  leads to subpar performance results. 
  In contrast, less complex (local) tagging models based on BiLSTMs perform robustly across classification scenarios, being able to catch long-range dependencies inherent to the AM problem. 
  Moreover, we find that jointly learning `natural' subtasks, in a 
  multi-task learning setup, improves performance. 
\end{abstract}

\section{Introduction}\label{sec:introduction}
Computational argumentation mining (AM) deals with finding argumentation structures in text. This involves several subtasks, such as: 
(a) separating argumentative units from non-argumentative units, also called `component segmentation';
(b) classifying argument components into classes such as ``Premise'' or ``Claim''; 
(c) finding relations between argument components;
(d) classifying relations into classes such as ``Support'' or ``Attack'' \cite{Persing:2016,Stab:2016}.

Thus, AM would have to detect claims and premises (reasons) in texts such as the following, where premise P supports claim C: 
\begin{quote}
 Since \underline{it killed many marine lives}$_\text{P}$ , \uwave{tourism has threatened nature}$_\text{C}$ .
\end{quote}
Argument structures in real texts are typically much more complex, cf.\ Figure \ref{fig:illustration}. 
\normalem

While different research has addressed different subsets of the AM problem (see below), the ultimate goal is to solve all of them, starting from unannotated plain text. Two recent approaches to this end-to-end learning scenario are \newcite{Persing:2016} and \newcite{Stab:2016}. 
Both solve the end-to-end task by first training independent models for each subtask and then defining an integer linear programming (ILP) model that encodes global constraints such as that each premise has 
a parent, etc. 
Besides their pipeline architecture the approaches also have in common that they heavily rely on hand-crafted features.   

Hand-crafted features pose a problem because AM is to some degree an ``arbitrary'' problem in that the notion of ``argument'' critically relies on the underlying argumentation theory \cite{Reed2008,Biran2011a,Habernal2015,Stab:2016}. 
Accordingly, datasets typically differ with respect to their 
annotation of (often rather complex) argument structure. 
Thus,
feature sets would have to be manually adapted to and designed for each new sample of data, a challenging task. 
The same critique applies to the designing of ILP constraints. 
Moreover, from a machine learning perspective, pipeline approaches are problematic 
because they solve subtasks independently and thus lead to error propagation rather than exploiting interrelationships between variables.  
In contrast to this, we investigate \emph{neural} techniques for end-to-end learning in computational AM, which do not require the hand-crafting of features or constraints.
The models we survey also all capture some notion of ``joint''---rather than ``pipeline''---learning. We investigate several approaches. 

First, we frame the end-to-end AM problem as a dependency parsing problem. Dependency parsing may be considered a natural choice for AM, because 
argument structures often form trees, or closely resemble them (see \S \ref{sec:data}). Hence, it is not surprising that `discourse parsing' \cite{Muller:2012} has been suggested for AM \cite{Peldszus:2015}. What distinguishes our approach from these previous ones is that we operate on the \emph{token} level, rather than on the level of components, because we address the end-to-end framework and, thus, do not assume that non-argumentative units have already been sorted out and/or that the boundaries of argumentative units are given. 

Second, we frame the problem as a sequence 
tagging problem. This is a natural choice especially for component identification (segmentation and classification), which is a typical entity recognition problem for which BIO tagging is a standard approach, pursued in AM, e.g., by \newcite{Habernal:2016}. The challenge in the end-to-end setting is to also include relations into the tagging scheme, which we realize by 
coding the distances between linked components into the tag label. 
Since related entities in AM are oftentimes several dozens of tokens apart from each other,  
neural sequence tagging models 
are in principle ideal candidates for such a 
framing because they can take into account \emph{long-range dependencies}---something that is inherently difficult to capture with traditional feature-based tagging models such as conditional random fields (CRFs). 

Third, we frame AM as a \emph{multi-task} (tagging) problem \cite{Caruana:1997,Collobert:2008}. We experiment with subtasks of AM---e.g., component identification---as auxiliary tasks and 
investigate whether this improves performance on the AM problem. Adding such subtasks can be seen as analogous to de-coupling, e.g., component identification from the full AM problem. 

Fourth, we evaluate the model of \newcite{Miwa:2016} that combines sequential (entity) and tree structure (relation) information and is in principle applicable to any problem where the aim is to extract entities and their relations. As such, this model makes fewer assumptions than our dependency parsing and tagging approaches. 

The contributions of this paper are as follows. 
(1) We present the \emph{first} \emph{neural} \emph{end-to-end} solutions to computational AM. 
(2) We show that several of them perform better than the state-of-the-art joint ILP model.
(3) We show that a framing of AM as a token-based dependency parsing problem is ineffective---in contrast to what has been proposed for systems that operate on the coarser {component} level and that
(4) a standard neural sequence tagging model that encodes distance information between components performs robustly in different environments.
Finally, (5) we show that a multi-task learning setup where natural subtasks of the full AM problem are added as auxiliary tasks improves performance.\footnote{Scripts that document how we ran our experiments are available from \url{https://github.com/UKPLab/acl2017-neural_end2end_AM}.} 


\section{Related Work} 
AM has applications in 
legal decision making \cite{Palau:2009,Moens:2007}, 
document summarization, and the analysis of scientific papers \cite{Kirschner2015}. 
Its importance for the educational domain has been highlighted by recent work on writing assistance \cite{ZhangL16} and essay scoring \cite{Persing:2015,Soma:2016}. 

Most works on AM address subtasks of AM such as locating/classifying components \cite{Florou:2013,Moens:2007,Rooney:2012,Burstein:2003,Levy:2014,Rinott:2015}.
Relatively few works address the full AM problem of component \emph{and} relation identification.
\newcite{Peldszus2016} present a corpus of microtexts containing only argumentatively relevant text of controlled complexity.
To our best knowledge, \newcite{Stab:2016} created the only corpus
of attested high quality which annotates the AM
problem in its entire complexity: it contains token-level annotations of
components, their types, 
as well as relations and their types. 

\section{Data}\label{sec:data}
We use the dataset of persuasive essays (PE) 
from \newcite{Stab:2016}, which contains student essays written in
response to controversial topics such as ``competition or cooperation---which is better?'' 

\begin{figure*}
  \begin{center}

\begin{tikzpicture}[->,>=stealth',shorten >=1pt,auto,node distance=2.5cm,
  thick,main node/.style={circle,fill=blue!20,draw,minimum width={width("C")+1pt},font=\sffamily\bfseries},none/.style={rectangle, fill=blue, minimum width=0.45cm, minimum height=0.45cm}]

  \begin{scope}
    \node[main node,fill=green,dashed] (MC1) {{\tiny MC$_1$}};
    \node[main node, fill=green,dashed] (MC2) [right=2.1cm of MC1] {{\tiny MC$_2$}};
    \node[main node,dotted] (CL1) [below left=0.8cm and 2.25cm of MC1] {{\tiny C$_1$}};
    \node[main node,dotted] (CL2) [below=0.55cm of MC1] {{\tiny C$_2$}};
    \node[main node,dotted] (CL3) [below right=0.8cm and 2.5cm of MC1] {{\tiny C$_3$}};
    \node[main node, fill=magenta, text=white] (P1) [below left=1.25cm of CL1] {{\tiny P$_1$}};
    \node[main node, fill=magenta, text=white] (P2) [below=0.6cm of CL1] {{\tiny P$_2$}};
    \node[main node, fill=magenta, text=white] (P3) [below right=1.25cm of CL1] {{\tiny P$_3$}};
    \node[main node, fill=magenta, text=white] (P4) [below=0.75cm of CL2] {{\tiny P$_4$}};
    \node[main node, fill=magenta, text=white] (P5) [below left=1.25cm of CL3] {{\tiny P$_5$}};
    \node[main node, fill=magenta, text=white] (P6) [below right=1.25cm of CL3] {{\tiny P$_6$}};


  \tikzset{Friend/.style   = {
                                 double          = black,
                                 double distance = 1pt}}
  \tikzset{Enemy/.style   = {
                                 double          = red,
                                 double distance = 1pt}}

  \draw[line width=0.75mm, draw=black](CL1) to (MC1); 
  \draw[line width=0.75mm, draw=gray,dashed](CL2) to (MC1);
  \draw[line width=0.75mm, draw=black](CL3) to (MC1);
  \draw[line width=0.75mm, draw=black](CL1) to (MC2); 
  \draw[line width=0.75mm, draw=gray,dashed](CL2) to (MC2);
  \draw[line width=0.75mm, draw=black](CL3) to (MC2);
  \draw[line width=0.75mm, draw=black](P1) to (CL1);
  \draw[line width=0.75mm, draw=black](P2) to (CL1);
  \draw[line width=0.75mm, draw=black](P3) to (CL1);
  \draw[line width=0.75mm, draw=black](P4) to (CL2);
  \draw[line width=0.75mm, draw=black](P5) to (CL3);
  \draw[line width=0.75mm, draw=black](P6) to (CL3);


  \end{scope}
  \begin{scope}[yshift=-3.75cm,xshift=-5cm]
    \node[none] (N0) {};
    \node[main node,fill=green,dashed] [right=0.2cm of N0] (MC1) {{\tiny MC$_1$}};
    \node[none] (N1) [right=0.2cm of MC1] {};
    \node[main node,dotted,thick] (CL1) [right=0.2cm of N1] {{\tiny C$_1$}};
    \node[none] (N2) [right=0.2cm of CL1] {};
    \node[main node, fill=magenta, text=white] (P1) [right=0.2cm of N2] {{\tiny P$_1$}};
    \node[none] (N3) [right=0.2cm of P1] {};
    \node[main node, fill=magenta, text=white] (P2) [right=0.2cm of N3] {{\tiny P$_2$}};
    \node[none] (N4) [right=0.2cm of P2] {};
    \node[main node, fill=magenta, text=white] (P3) [right=0.2cm of N4] {{\tiny P$_3$}};
    \node[none] (N5) [right=0.2cm of P3] {};
    \node[main node, fill=magenta, text=white] (P4) [right=0.2cm of N5] {{\tiny P$_4$}};
    \node[none] (N6) [right=0.2cm of P4] {};
    \node[main node,dotted,thick] (CL2) [right=0.2cm of N6] {{\tiny C$_2$}};
    
    \node[none] (N7) [below=0.5cm of N0] {};
    \node[main node, fill=magenta, text=white] (P5) [right=0.2cm of N7] {{\tiny P$_5$}};
    \node[none] (N8) [right=0.2cm of P5] {};
    \node[main node, fill=magenta, text=white] (P6) [right=0.2cm of N8] {{\tiny P$_6$}};
    \node[none] (N9) [right=0.2cm of P6] {};
    \node[main node,thick,dotted] (CL3) [right=0.2cm of N9] {{\tiny C$_3$}};
    \node[none] (N10) [right=0.2cm of CL3] {};
    \node[main node, fill=green,dashed] (MC2) [right=0.2cm of N10] {{\tiny MC$_2$}};
    \node[none] (N11) [right=0.2cm of MC2] {};

  \tikzset{Friend/.style   = {
                                 double          = black,
                                 double distance = 1pt}}
  \tikzset{Enemy/.style   = {
                                 double          = red,
                                 double distance = 1pt}}


\end{scope}
\end{tikzpicture}
  \end{center}
  \caption{ Bottom: Linear argumentation structure in a student essay. The essay is comprised of non-argumentative units (square) and argumentative units of different types: Premises (P), claims (C) and major claims (MC).
  Top: Relationsships between argumentative units. Solid arrows are support (for), dashed arrows are attack (against).}
  \label{fig:illustration}
\end{figure*}
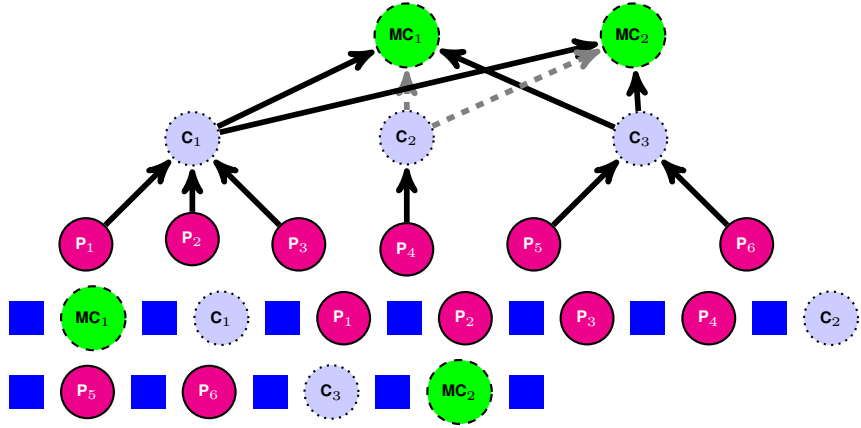
\begin{table}[!htb]
  \centering
  \small
  \begin{tabular}{l|rr} \toprule
  & Train & Test \\ \midrule
  Essays & 322 & 80 \\
  Paragraphs & 1786 & 449 \\
  Tokens & 118648 & 29538 \\
  \bottomrule
  \end{tabular}
  \caption{Corpus statistics}
  \label{table:cstats}
\end{table}

As Table \ref{table:cstats} details, 
the corpus consists of 402 essays, 80 of which are reserved for
testing. The annotation distinguishes between \textbf{major claims}
(the central position of an author with respect to the essay's
topic), \textbf{claims} (controversial statements that are
either \emph{for} or \emph{against} the major claims),
and \textbf{premises}, which give reasons for claims or other premises
and either \emph{support} or \emph{attack} them. 
Overall, there are 751 major claims, 1506 claims, and 3832
premises. There are 5338 relations, most of which are supporting relations ($>$90\%). 

The corpus has a special structure, illustrated in
Figure \ref{fig:illustration}. First, major claims relate to no other
components. Second, claims always relate to all other major
claims.\footnote{All MCs are considered as equivalent in meaning.}
Third, each premise relates to exactly one claim or premise.  
Thus, the argument structure in each essay is---almost---a tree.
Since there may be several major claims, each claim potentially connects
to multiple targets, violating the tree structure. This
poses no problem, however, since we can ``loss-lessly'' re-link the
claims to one of the major claims (e.g., the last major claim in
a document) and create a special root node to which the major claims
link. From this tree, the actual graph can be uniquely reconstructed. 

There is another peculiarity of this data. Each 
essay 
is divided into paragraphs, of which there are 2235 in total. 
The argumentation structure is completely contained within a paragraph,
except, possibly, 
for the relation from claims to major claims. Paragraphs have an
average length of 66 tokens and are therefore much shorter than
essays, which have an average length of 368 tokens. Thus, prediction
on the paragraph level is easier than prediction on the essay level,
because there are fewer components in a paragraph and hence
fewer possibilities of source and target components in argument
relations. 
The same is true for component classification: a paragraph can never contain premises only, for example, since premises link to other components.

\section{Models}\label{sec:models}

This section describes our neural network framings for 
end-to-end AM.

\paragraph{Sequence Tagging}
is the problem of
assigning each element in a stream of input tokens a label.
In a neural 
context, the natural choice for 
tagging problems are
recurrent neural nets (RNNs) in which a hidden vector representation
$\mathbf{h}_t$ at time point $t$ depends on the previous hidden vector
representation $\mathbf{h}_{t-1}$ and the input $\mathbf{x}_t$. In
this way, an infinite window (``long-range dependencies'') around the
current input token 
$\mathbf{x}_t$ can be taken into account when making an output
prediction $\mathbf{y}_t$.
We choose particular RNNs, namely, LSTMs \cite{Hochreiter:1997}, which are popular for being able to address vanishing/exploding gradients
problems. 
In addition to considering a
left-to-right flow of information, bidirectional LSTMs (BL) also
capture information to the right of the current input token. 

The most recent generation of neural tagging models add label
dependencies to BLs, so that successive output decisions are not made independently. 
This class of models is called BiLSTM-CRF (BLC)
\cite{Huang:2015}.
The model
of \newcite{Ma:2016} adds convolutional neural nets (CNNs) on the
character-level to BiLSTM-CRFs, leading to BiLSTM-CRF-CNN (BLCC) models. The
character-level CNN may address problems of out-of-vocabulary words,
that is, words not seen during training.
%

\textbf{AM as Sequence Tagging}: 
We frame AM as the following sequence tagging 
problem. Each input token has an associated label from $\mathcal{Y}$,
where 
\begin{equation}\label{eq:labelset}
\small
  \begin{split}
  \mathcal{Y} = \{(b,t,d,s)\,|\,& b\in\{\text{B},\text{I},\text{O}\},
    t\in\{\text{P},\text{C},\text{{MC}},\perp\}, \\
    & d\in\{\ldots,-2,-1,1,2,\ldots,\perp\},\\
    & s\in \{\text{Supp},\text{Att},\text{For},\text{Ag},\perp\}\}.
  \end{split}
\end{equation}
In other words, $\mathcal{Y}$ consists of all four-tuples $(b,t,d,s)$ where $b$ is a BIO encoding indicating whether the current token is non-argumentative (O) or begins (B) or continues (I) a component; $t$ indicates the \emph{type} of the 
component (claim C, premise P, or major claim $\text{{MC}}$ for our data).
Moreover, $d$ encodes the distance---measured in number of components---between the current component and the component it relates to. We encode the same $d$ value for each token in a given component. Finally, $s$ is the relation type (``stance'') between two components and its value may be Support (Supp), Attack (Att), or For or Against (Ag). We also have a special symbol $\perp$ that indicates when a particular slot is not filled: e.g., a non-argumentative unit ($b=$ O) has neither component type, nor relation, nor relation type. We refer to this framing as \texttt{STag$_T$} (for ``\emph{S}imple \emph{Tag}ging''), where $T$ refers to the tagger used. For the example from \S\ref{sec:introduction}, 
our coding would hence be: 
\begin{table}[!htb]
  \centering
  {\small 
  \begin{tabular}{llllll}
    Since & it & killed & many \\ \vspace{.2cm}
    (O,$\perp$,$\perp$,$\perp$) & (B,P,1,Supp) & (I,P,1,Supp) & (I,P,1,Supp)
    \\  
    marine & lives & , & tourism \\ \vspace{.2cm}
    (I,P,1,Supp) & (I,P,1,Supp) & (O,$\perp$,$\perp$,$\perp$) & (B,C,$\perp$,For)\\  
    has & threatened & nature & . \\  
    (I,C,$\perp$,For) & (I,C,$\perp$,For) & (I,C,$\perp$,For) & (O,$\perp$, $\perp$, $\perp$) \\
  \end{tabular}
  }
\end{table}

\begin{figure}[!htb]
  \centering
  \scalebox{.6}
  {\input{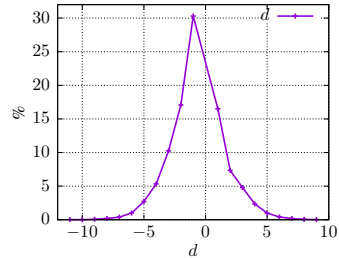}}
  \caption{Distribution of distances $d$ between components in PE dataset.}
  \label{fig:distribution}
\end{figure}

While the size of the label set $\mathcal{Y}$ is potentially infinite,
we would expect it to be finite even in a potentially infinitely large data set, because humans also have only finite memory and are therefore expected to keep related components close in textual space. Indeed, as Figure \ref{fig:distribution} shows, in our PE essay data set about 30\% of all relations between components have distance $-1$, that is, they follow the claim or premise that they attach to. Overall, 
around 2/3 of all relation distances $d$ lie in $\{-2,-1,1\}$. However, the figure also illustrates that there are  indeed long-range dependencies: distance values between $-11$ and $+10$ are observed in the data. 

 \paragraph{Multi-Task Learning}        
Recently, there has been a lot of interest in so-called multi-task learning (MTL) scenarios, where several tasks are learned \emph{jointly} \cite{Sogaard:2016,Peng:2016,Yang:2016,Rusu:2016,Hector:2017}. It has been argued that such learning scenarios are closer to human learning because humans often transfer knowledge between several domains/tasks. In a neural 
 context, MTL is typically implemented via weight sharing: several tasks are trained in the same network architecture, thereby sharing a substantial portion of network's parameters. This forces the network to learn \emph{generalized} representations. 

In the MTL framework of \newcite{Sogaard:2016} 
the underlying model is a BiLSTM with several hidden layers. Then, given different tasks, each task $k$ `feeds' from one of the hidden layers in the network. In particular, the hidden states encoded in a specific layer are fed into a multiclass classifier $f_k$.  
The same work has demonstrated that this MTL protocol may be successful when there is a hierarchy between tasks and `lower' tasks feed from lower layers. 

\textbf{AM as MTL}: We use the same framework \texttt{STag$_{T}$} for modeling AM as MTL. However, we in addition train auxiliary tasks in the 
network---each with a distinct label set $\mathcal{Y}'$. 
\paragraph{Dependency Parsing}
methods can be classified into graph-based and 
transition-based approaches \cite{Kiperwasser:2016}. \emph{Transition-based} parsers
encode the parsing problem as a sequence of configurations which may
be modified by application of {actions} such as shift, reduce,
etc. The system starts with an initial configuration in which 
sentence elements are on a \emph{buffer} and a \emph{stack}, and a
classifier successively decides which action to take next, leading to
different configurations. The system terminates after a 
finite number 
of actions, and the parse tree is read off the terminal
configuration. \emph{Graph-based} parsers solve a structured prediction
problem in which the goal is learning a scoring function over
dependency trees such that correct trees are scored above all
others. 


Traditional 
dependency parsers used hand-crafted
feature functions that look at ``core'' elements such as ``word on top
of the stack'', ``POS of word on top of the stack'', 
and conjunctions of 
core
features such as ``word is X and POS is Y'' (see
\newcite{McDonald:2005}). Most neural 
parsers have not entirely abandoned feature engineering. Instead, they
rely, for example, on encoding the core features of parsers as
low-dimensional embedding vectors \cite{Chen:2014} but ignore feature
combinations. \newcite{Kiperwasser:2016} design a neural 
parser that uses only four features: the BiLSTM vector representations
of the top 3 items on the stack and the first item on the
buffer. In contrast, \newcite{Dyer:2015}'s 
neural 
parser associates 
each stack with a ``stack LSTM'' that encodes their 
contents. Actions are chosen based on the stack LSTM representations
of the stacks, and no more feature engineering is necessary.   
Moreover, their parser has thus access to any part of the input, its history and stack contents.  

\begin{figure}[!htb]
  \centering
  \scalebox{1}{
  \begin{tikzpicture}[->,>=stealth',shorten >=1pt,auto,node distance=.3cm,
  thick,main node/.style={circle,fill=white!20,draw,minimum width={width("1")+1pt},font=\sffamily\bfseries},none/.style={rectangle, fill=blue, minimum width=0.1cm, minimum height=0.1cm, text=white}]

    \node[none] (1) {\tiny{1}};
    \node[main node] (2) [right=0.08cm of 1,fill=magenta, text=white] {\tiny{2}};
    \node[main node] (3) [right=0.08cm of 2,fill=magenta, text=white] {\tiny{3}};
    \node[main node] (4) [right=0.08cm of 3,fill=magenta, text=white] {\tiny{4}};
    \node[main node] (5) [right=0.08cm of 4,fill=magenta, text=white] {\tiny{5}};
    \node[main node,fill=magenta, text=white] (6) [right=0.08cm of 5] {\tiny{6}};
    \node[none] (7) [right=0.08cm of 6] {\tiny{7}};
    \node[main node,fill=blue!20,dotted] (8) [right=0.08cm of 7] {\tiny{8}};
    \node[main node,fill=blue!20,dotted] (9) [right=0.08cm of 8] {\tiny{9}};
    \node[main node,fill=blue!20,dotted] (10) [right=0.08cm of 9] {\scalebox{.375}{10}};
    \node[main node,fill=blue!20,thick,dotted] (11) [right=0.08cm of 10] {\scalebox{.375}{11}};
    \node[none] (12) [right=0.08cm of 11] {\scalebox{.375}{12}};

  \tikzset{Friend/.style   = {
                                 double          = black,
                                 double distance = 1pt}}
  \tikzset{Enemy/.style   = {
                                 double          = red,
                                 double distance = 1pt}}

  \path (1) edge[bend right=50] node [midway,below,fill=white] {\tiny{O}} (12);
  \path (2) edge[bend right=60] node [below,right,fill=white] {\tiny{(\text{B},\text{P},\text{Supp})}} (8);
  \path (3) edge[bend right=49] node[right,fill=white] {\tiny{(\text{I},\text{P},\text{Supp})}} (8);
  \path (4) edge[bend right=50] node [left] {} (8);
  \path (5) edge[bend right=50] node [left] {} (8);
  \path (6) edge[bend right=50] node [left] {} (8);
  \path (7) edge[bend right=-50] node [above] {\tiny{O}} (12);
  \path (8) edge[bend right=-40] node [left,fill=white] {\tiny{(\text{B},\text{C},\text{For})}} (12);
  \path (9) edge[bend right=-40] node [] {} (12); 
  \path (10) edge[bend right=-40] node [left] {} (12); 
  \path (11) edge[bend right=-40] node [left] {} (12);
\end{tikzpicture}}
  \caption{
    Dependency representation of sample sentence from \S\ref{sec:introduction}. Links and selected labels. 
    }
  \label{pic:example}
\end{figure}
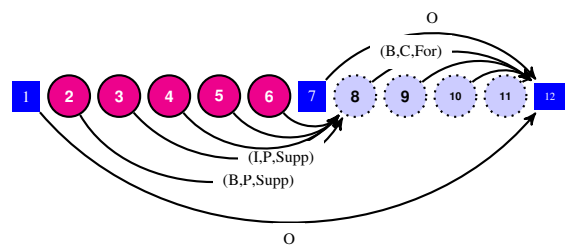

\textbf{AM as Dependency Parsing}:
To frame a problem as a dependency parsing problem, each
instance of the problem must be encoded as a directed tree, where
tokens have heads, which in turn are labeled. For end-to-end AM, we propose the
framing illustrated in Figure \ref{pic:example}. 
We highlight two design decisions, the remaining are analogous and/or can be read off the figure. 
\begin{itemize}
\itemsep0em
\item The head of each non-argumentative text token is
  the document terminating token \texttt{END}, 
which is a punctuation mark in all our cases. The label of this link is \text{O}, the symbol for non-argumentative units. 
\item The head of each token in a premise is the \emph{first} token of the claim or premise that it links to. The label of each of these links is 
$(b,\text{P},\text{Supp})$ or $(b,\text{P},\text{Att})$
depending on whether a premise ``supports'' or ``attacks'' a claim or premise; $b\in\{\text{B},\text{I}\}$. 
\end{itemize}




\paragraph{LSTM-ER}
\newcite{Miwa:2016} present a neural end-to-end system for identifying both entities as well as relations between them.
Their entity detection system is a BLC-type tagger and their relation detection system is a neural net that predicts a relation for each pair of detected entities. This relation module is a TreeLSTM model that makes use of dependency tree information. 
In addition to de-coupling entity and relation detection but {jointly} modeling them,\footnote{By `de-coupling', we mean that both tasks are treated separately rather than merging entity and relation information in the same tag label (output space).
Still, a joint model like that of \newcite{Miwa:2016} de-couples the two tasks in such a way that many model parameters are shared across the tasks, similarly as in MTL.}
pretraining on entities and scheduled sampling \cite{Bengio2015} is applied to prevent low performance at early training stages of entity detection and relation classification. 
To adapt LSTM-ER for the argument structure encoded in the PE dataset,  
we model three types of entities (premise, claim, major claim) and four types of relations (for, against, support, attack). 

We use the \textbf{feature-based ILP model} from \newcite{Stab:2016} as a comparison system. This system solves the subtasks of AM---component segmentation, component classification, relation detection and classification---independently. Afterwards, it defines an  ILP model with various constraints to enforce valid argumentation structure. As features it uses structural, lexical, syntactic and context features, cf.\ \newcite{Stab:2016} and \newcite{Persing:2016}.  

Summarizing, 
we distinguish our framings in terms of \emph{modularity} and in terms of their \emph{constraints}. \emph{Modularity}: Our dependency parsing framing and LSTM-ER are more modular than \texttt{STag$_{T}$} because they de-couple relation information from entity information. However, (part of) this modularity can be regained by using \texttt{STag$_{T}$} in an MTL setting. Moreover, since entity and relation information are considerably different, 
such a de-coupling may be advantageous. 
\emph{Constraints}: LSTM-ER
can, in principle, model any kind of---even many-to-many---relationships between detected entities. Thus, 
it is not guaranteed to produce trees, as we observe in AM datasets. \texttt{STag$_T$} also does not need to produce trees, but it more severely restricts search space than does LSTM-ER: each token/component can only relate to one (and not several) other tokens/components. The same constraint is enforced by the dependency parsing framing.
All of the tagging modelings, including LSTM-ER, are \emph{local} models whereas our parsing framing is a \emph{global} model: it globally enforces a tree structure on the token-level. 

Further remarks: (1) part of the TreeLSTM modeling inherent to LSTM-ER is ineffective for our data because this modeling exploits 
dependency tree structures on the \emph{sentence} level, while relationships between components are almost never on the sentence level. In our data, 
roughly 92\% of all relationships are between components that appear in different sentences. Secondly, (2) that a model \emph{enforces} a constraint does not necessarily mean that it is more suitable for a respective task. It has frequently been observed 
that models tend to produce output consistent with constraints in their training data in such situations \cite{Zhang:2016,Hector:2017}; 
thus, they have \emph{learned} the constraints.

\section{Experiments}\label{sec:experiments}
This section presents and discusses the empirical results for the AM framings outlined in \S\ref{sec:models}.  
We relegate issues of \emph{pre-trained word embeddings}, \emph{hyperparameter optimization} and further \emph{practical issues} to the supplementary material. Links to software used
as well as some additional error analysis
can also be found there. 

\paragraph{Evaluation Metric} We adopt the evaluation metric suggested
in \newcite{Persing:2016}. This computes true positives TP, false
positives FP, and false negatives FN, and from these calculates
component and relation $F_1$ scores as $F_1
= \frac{2\text{TP}}{2\text{TP}+\text{FP}+\text{FN}}$.
For space reasons, we refer to \newcite{Persing:2016} for specifics, but to illustrate, 
for \emph{components}, true positives are defined as the set of 
components in the gold standard for which there exists a predicted
component with the same type that `matches'. \newcite{Persing:2016} define a
notion of what we may term `level $\alpha$ matching': for example,
at the 100\% 
level (exact match) predicted and gold components must have
exactly the same 
spans, whereas at the 50\% level 
they must only share at 
least 50\% of their tokens (approximate match).
We refer to these scores as C-F1 (100\%) and C-F1 (50\%), respectively. For \emph{relations}, an analogous F1 score is determined, which we denote by R-F1 (100\%) and R-F1 (50\%).
We note that R-F1 scores depend on C-F1 scores because 
correct relations must have correct arguments. 
We also define a `global' F1 score, which is the F1-score of C-F1 and R-F1.

Most of our results are shown in Table \ref{table:parsing}.

\paragraph{(a) Dependency Parsing} We show results for the two feature-based
parsers MST \cite{McDonald:2005}, Mate \cite{Bohnet:2012} as well as
the neural parsers by \newcite{Dyer:2015} (LSTM-Parser) and 
\newcite{Kiperwasser:2016} (Kiperwasser).
We train and test
all parsers on the paragraph level, because training them on essay 
level was typically too memory-exhaustive.

MST mostly labels only non-argumentative units correctly, except 
for 
recognizing individual major claims, but never finds 
their exact spans (e.g., \emph{``tourism can create negative
impacts on''} while the gold major claim is \emph{``international tourism can
create negative impacts on the destination countries''}). Mate is
slightly better and in particular recognizes several major claims
correctly. 
Kiperwasser 
performs decently on the approximate match level, but not on exact level. Upon inspection,
we find that the parser often predicts `too large' 
component spans, e.g., by including following punctuation. 
The best parser by far is the
LSTM-Parser. It is over 100\% better than Kiperwasser 
on exact spans and still several
percentage points on approximate spans. 
\begin{table*}[!htb]
  \centering
  {\scriptsize
  \begin{tabular}{l|r|rr|rr|rr||r|rr|rr|rr} \toprule
  & \multicolumn{7}{c||}{Paragraph level} & \multicolumn{7}{c}{Essay level} \\  \midrule             
    & Acc. & \multicolumn{2}{c}{C-F1} & \multicolumn{2}{c}{R-F1} & \multicolumn{2}{c}{F1} & Acc. & \multicolumn{2}{c}{C-F1} & \multicolumn{2}{c}{R-F1} & \multicolumn{2}{c}{F1} \\
    & & 100\% & 50\% & 100\% & 50\% & 100\% & 50\% & & 100\% & 50\% & 100\% & 50\% & 100\% & 50\%\\
     \midrule  
    {\small MST-Parser} & 31.23 & 0 & 6.90 & 0 & 1.29 & 0 & 2.17 & & & & & & &\\
    {\small Mate} & 22.71 & 2.72 & 12.34 & 2.03 & 4.59 & 2.32 & 6.69 & & & & & & &\\
    {\small Kiperwasser} & 52.80 & 26.65 & 61.57 & 15.57 & 34.25 & 19.65 & 44.01 & & & & & & & \\
     {\small LSTM-Parser} & 55.68 & 58.86 & 68.20 & 35.63 & 40.87 & 44.38 & 51.11 & & & & & & & \\
    \midrule
    \texttt{STag$_{\text{BLCC}}$} & 59.34 & 66.69 & 74.08 & 39.83 & 44.02 & 49.87 & 55.22 & \textbf{60.46} & 63.23 & 69.49 & \textbf{34.82} & \textbf{39.68} & \textbf{44.90} & \textbf{50.51} \\ \midrule
    LSTM-ER & \textbf{61.67} & \textbf{70.83} & \textbf{77.19} & \textbf{45.52} & \textbf{50.05} & \textbf{55.42} & \textbf{60.72} & 54.17 & \textbf{66.21} & \textbf{73.02} & 29.56 & 32.72 & 40.87 & 45.19 \\ \midrule
    ILP & 60.32 & 62.61 & 73.35 & 34.74 & 44.29 & 44.68 & 55.23 & & & & & & & \\
    \bottomrule
  \end{tabular}
  }
  \caption{
   Performance of dependency parsers, \texttt{STag$_{\text{BLCC}}$}, LSTM-ER and ILP (from top to bottom). 
   The ILP model operates on both levels. Best scores in each column in bold (signific.\ at $p<0.01$; Two-sided Wilcoxon signed rank test, pairing F1 scores for documents). We also report token level accuracy.
  }
  \label{table:parsing}
\end{table*}

How does performance change when we switch to the essay level?
For the LSTM-Parser, the best performance on essay level is 32.84\%/47.44\%  C-F1 
(100\%/50\% level), and 9.11\%/14.45\% on R-F1, but performance result
varied drastically between different parametrizations. Thus, the performance
drop between paragraph and essay level is in any case immense. 

Since the employed features of modern feature-based parsers are rather
general---such as distance 
between words or word identities---
we had expected them to perform much 
better.
The minimal feature set employed by 
Kiperwasser 
is apparently not sufficient for accurate
AM but still a lot more powerful than the
hand-crafted feature approaches.
We
hypothesize that the LSTM-Parser's good performance, relative to the other parsers,  is due
to its encoding of the \emph{whole} 
stack history---rather than just the top elements on the stack as in
Kiperwasser--- 
which makes it aware of much larger
`contexts'. While the drop in performance from paragraph to 
essay level is expected, the LSTM-Parser's deterioration 
is much more severe 
than the other models' surveyed below. 
We believe that this is due to
a mixture of 
the following: (1) `capacity', i.e., model complexity, of the parsers---that is, risk of overfitting; and (2) few, but very long sequences on essay level---that is, little training data (trees), paired with a huge search space on each train/test instance, namely, the number of possible trees on $n$ tokens. See also our discussions below, particularly, our stability analysis. 

\paragraph{(b) Sequence Tagging}
For these experiments, we use the BLCC tagger from \newcite{Ma:2016} and refer to the resulting system as \texttt{STag$_{\text{BLCC}}$}. Again, we observe that paragraph level is considerably easier than essay level; e.g., for relations,   
there is $\sim$5\% points increase from essay to paragraph level. Overall, \texttt{STag$_{\text{BLCC}}$} is $\sim$13\% better than the best parser for C-F1 and $\sim$11\% better for R-F1 on the paragraph level. 
Our explanation is 
that taggers are simpler local models, and thus need less training data and are less prone to overfitting. Moreover, they can much better deal with the long sequences 
because
they 
are 
largely invariant to length: e.g., it does in principle not matter, from a parameter estimation perspective, whether we train our taggers on two sequences of lengths $n$ and $m$, respectively, or on one long sequence of length $n+m$. 


\paragraph{(c) MTL}
As indicated, we use the MTL tagging framework from \newcite{Sogaard:2016} for multi-task experiments. The underlying tagging framework is weaker than that of BLCC: there is no CNN which can take subword information into account and there are no dependencies between output labels: each tagging prediction is made independently of the other predictions. 
We refer to this system as \texttt{STag$_\text{BL}$}.

Accordingly, as Table \ref{table:mtl} shows for the essay level (paragraph level omitted for space reasons), results are generally weaker: For exact match, C-F1 values are about $\sim$10\% points below those of \texttt{STag$_\text{BLCC}$}, while approximate match performances are much closer. Hence, the independence assumptions of the BL tagger apparently lead to more `local' errors such as exact argument span identification (cf.\ error analysis). 
An analogous trend holds for argument relations.

\emph{Additional Tasks:} We find that when we train \texttt{STag$_\text{BL}$} with only its main task---with label set $\mathcal{Y}$ as in Eq.~\eqref{eq:labelset}---the overall result is worst. In contrast, when we include the `natural subtasks' ``C'' (label set $\mathcal{Y}_C$ consists of the projection on the coordinates $(b,t)$ in $\mathcal{Y}$) and/or ``R'' (label set $\mathcal{Y}_R$ consists of the projection on the coordinates $(d,s)$), performance increases typically by a few percentage points. This indicates that complex sequence tagging may benefit when we train a ``sub-labeler'' in the same neural architecture, a finding that may be particularly relevant for morphological POS tagging \cite{Mueller:2013}. Unlike \newcite{Sogaard:2016}, we do not find that the optimal architecture is the one in which ``lower'' tasks (such as C or R) feed from lower layers. In fact, in one of the best parametrizations 
the C task and the full task feed from the same layer in the deep BiLSTM. Moreover, we find that the C task is consistently more helpful as an auxiliary task than the R task.
\begin{table}[!htb]
  \centering
  {\scriptsize
    \setlength{\tabcolsep}{4.5pt} 
   \begin{tabular}{p{0.11\textwidth}|rr|rr|rr} \toprule
    & \multicolumn{2}{c}{C-F1} & \multicolumn{2}{c}{R-F1} & \multicolumn{2}{c}{F1} \\ 
    & 100\% & 50\% & 100\% & 50\% & 100\% & 50\% \\ \midrule
    $\mathcal{Y}$-3 & 49.59 & 65.37 & 26.28 & 37.00 & 34.35 & 47.25 \\ \midrule
    $\mathcal{Y}$-3:$\mathcal{Y}_C$-1 & 54.71 & 66.84 & 28.44 & 37.35 & 37.40 &  47.92\\
    $\mathcal{Y}$-3:$\mathcal{Y}_R$-1 & 51.32 & 66.49 & 26.92 & 37.18 & 35.31 & 47.69 \\
    $\mathcal{Y}$-3:$\mathcal{Y}_C$-3 & \textbf{54.58} & 67.66 & \textbf{30.22} & \textbf{40.30} & \textbf{38.90} & \textbf{50.51} \\
    $\mathcal{Y}$-3:$\mathcal{Y}_R$-3 & 53.31 & 66.71 & 26.65 & 35.86 & 35.53 & 46.64 \\
    $\mathcal{Y}$-3:$\mathcal{Y}_C$-1:$\mathcal{Y}_R$-2 & 52.95 & \textbf{67.84} & 27.90 & 39.71 & 36.54 & 50.09 \\
    $\mathcal{Y}$-3:$\mathcal{Y}_C$-3:$\mathcal{Y}_R$-3 & 54.55 & 67.60 & 28.30 & 38.26 & 37.26 & 48.86 \\
    \bottomrule
  \end{tabular}
  }
  \caption{ Performance of MTL sequence tagging approaches, essay level. Tasks separated by ``:''. Layers from which tasks feed are indicated by respective numbers.}
  \label{table:mtl}
\end{table}

On essay level, \textbf{(d) LSTM-ER} performs very well on component 
identification (+5\% C-F1 compared to \texttt{STag$_{\text{BLCC}}$}), but rather poor on relation identification (-18\% R-F1). Hence, its overall F1 on essay level is considerably below that of \texttt{STag$_{\texttt{BLCC}}$}.
In contrast, LSTM-ER trained and tested on paragraph level substantially outperforms all other systems discussed, 
both for component 
as well as for relation identification. 

We think that its generally excellent performance on components is due to LSTM-ER's de-coupling of component and relation tasks. Our findings indicate that a similar result can be achieved for \texttt{STag$_{T}$} via MTL when components and relations are included as auxiliary tasks, cf.~Table \ref{table:mtl}. For example, the 
improvement of LSTM-ER over \texttt{STag$_{\text{BLCC}}$}, for C-F1, roughly matches the increase for \texttt{STag$_{\text{BL}}$} when including components and relations separately 
($\mathcal{Y}\text{-3}\text{:}\mathcal{Y}_{C}\text{-3}\text{:}\mathcal{Y}_R\text{-3}$) over not including them as auxiliary tasks ($\mathcal{Y}\text{-3}$).
Lastly, the better performance of LSTM-ER over
\texttt{STag$_{\text{BLCC}}$} 
for relations on paragraph level appears to be a consequence of its better performance on components. E.g., when both arguments are correctly predicted, \texttt{STag$_{\text{BLCC}}$} has even higher chance of getting their relation correct than LSTM-ER (95.34\% vs. 94.17\%).

\begin{figure}
  \centering
  \scalebox{0.6}{
  \input{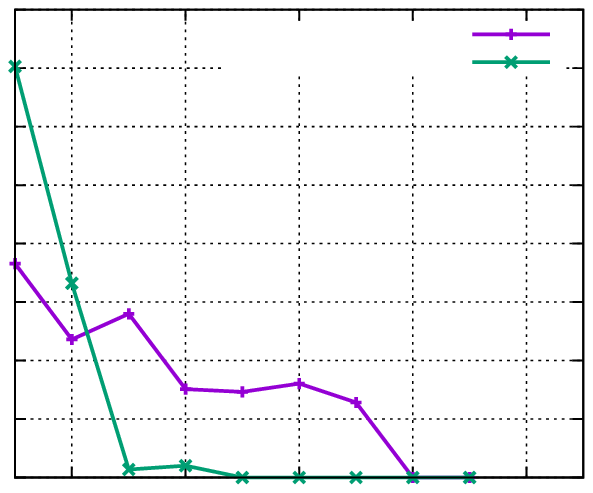}}
  \caption{Probability of correct relation identification given true distance is $|d|$.}
  \label{fig:distances}
\end{figure}

Why does LSTM-ER degrade so much on essay level for R-F1? As said, text sequences are much longer on essay level than on paragraph level---hence, there are on average many more entities on essay level.
Thus, there are also many more 
possible relations between all entities discovered in a text---namely, there are $O(2^{m^2})$ possible relations between $m$ discovered components. Due to its generality, LSTM-ER considers all these relations as plausible, while \texttt{STag$_T$} does not (for any of choice of $T$): e.g., our coding explicitly constrains each premise to link to exactly \emph{one} other component, rather than to $0,\ldots,m$ possible components, as LSTM-ER allows. In addition, our explicit coding of distance values $d$ biases the learner $T$ to reflect the distribution of distance values found in real essays---namely, that related components are typically close in terms of the number of components between them. In contrast, LSTM-ER only mildly prefers 
short-range dependencies over long-range dependencies, cf.\ Figure \ref{fig:distances}. 

The \textbf{(e) ILP} has access to both paragraph and essay level information and thus has always more information than all neural systems compared to. 
Thus, it also knows in which paragraph in an essay it  is. This is useful particularly for major claims, which always occur in first or last paragraphs in our data. Still, its performance is equal to or lower than that of LSTM-ER and 
\texttt{STag$_{\text{BLCC}}$} 
when both 
are evaluated on paragraph level.   

\subsection*{Stability Analysis}
Table \ref{table:stability} shows averages and standard deviations of two selected models, namely, the \texttt{STag$_{\text{BLCC}}$} tagging framework as well as the LSTM-Parser over several different runs (different random initializations as well as different hyperparameters as discussed in the supplementary material). These results detail that the taggers have lower standard deviations than the parsers. The difference is particularly striking on the essay level where the parsers often completely fail to learn, that is, their performance scores are close to 0\%. As discussed above, we attribute this to the parsers' increased model capacity relative to the taggers, which makes them more prone to overfitting. Data scarcity is another very likely source of error in this context, as the parsers only observe 322 (though very rich) trees in the training data, while the taggers are always roughly trained on 120K tokens. On paragraph level, they do observe more trees, namely, 1786. 
\begin{table}[!htb]
  \centering
  \small
  \begin{tabular}{l|ll} \toprule
        &  \texttt{STag$_{\text{BLCC}}$} & LSTM-Parser \\ \midrule
  Essay & 60.62$\pm$3.54 & 9.40$\pm$13.57 \\
  Paragraph & 64.74$\pm$1.97 & 56.24$\pm$2.87\\ \bottomrule
  \end{tabular}
  \caption{C-F1 (100\%) in \% for the two indicated systems; essay vs.\ paragraph level. Note that the mean performances are lower than the majority performances over the runs given in Table \ref{table:parsing}.}
  \label{table:stability}
\end{table}

\section*{Error analysis}
A systematic source of errors for all systems is detecting exact argument spans (segmentation). For instance, the ILP system predicts the following premise: \emph{``As a practical epitome , students should be prepared to present in society after their graduation''}, while the gold premise omits the preceding discourse marker, and hence reads: \emph{``students should be prepared to present in society after their graduation''}. On the one hand, it has been observed that even humans have problems exactly identifying such entity boundaries \cite{Persing:2016,Yang:2013}. On the other hand, our results in Table \ref{table:parsing} indicate that the neural taggers BLCC and BLC (in the LSTM-ER model) are much better at such exact identification than either the ILP model or the neural parsers.
While the parsers' problems are most likely due to model complexity,
we hypothesize that the ILP model's increased error rates stem from a weaker underlying tagging model (feature-based CRF vs.\ BiLSTM) and/or its features.\footnote{The BIO tagging task is independent and thus not affected by the ILP constraints in the model of \newcite{Stab:2016}. The same holds true for the model of \newcite{Persing:2016}.} In fact, as Table \ref{table:BIO} shows, the macro-F1 scores\footnote{Denoted $\textrm{\emph{Fscore}}_M$ in \newcite{Sokolova.Lapalme.2009}.} on \emph{only} the component segmentation tasks (BIO labeling) are substantially higher for both LSTM-ER and \texttt{STag}$_{\text{BLCC}}$ than for the ILP model.
Noteworthy, the two neural systems even outperform the human upper bound (HUB) in this context, 
reported as 88.6\% in \newcite{Stab:2016}. 

\begin{table}[!htb]
\centering
\small
\begin{tabular}{l|ccc|c}
  \toprule
        & \texttt{STag}$_{\text{BLCC}}$ & LSTM-ER & ILP & HUB \\ \midrule
  Essay & 90.04 & 90.57 & & \\
  Paragraph & 88.32 & 90.84 & 86.67 & 88.60 \\ \bottomrule
\end{tabular}
\caption{F1 scores in \% on BIO tagging task.}
\label{table:BIO}
\end{table}



\section{Conclusion}\label{sec:conclusion}
We present the first study on neural end-to-end AM. We experimented with different \emph{framings}, such as encoding AM as a dependency parsing problem, as a sequence tagging problem with particular label set, as a multi-task sequence tagging problem, and as a problem with both sequential and tree structure information. 
We show that (1) neural computational AM is as good or (substantially) better than a competing feature-based ILP formulation, while eliminating the need for manual feature engineering and costly ILP constraint designing. 
(2) BiLSTM taggers perform very well for component identification, as demonstrated for our \texttt{STag}$_T$ frameworks, for $T=$ BLCC and $T=$ BL, as well as for LSTM-ER (BLC tagger).
(3) (Naively) coupling component and relation identification is not optimal, but both tasks should be treated separately, but modeled jointly. 
(4) Relation identification is more difficult: when there are few entities in a text (``short documents''), a more general framework such as that provided in LSTM-ER performs reasonably well. When there are many entities (``long documents''), a more restrained modeling is preferable. These are also our \emph{policy recommendations}. 
Our work yields new state-of-the-art results in end-to-end AM on the PE dataset from \newcite{Stab:2016}. 

Another possible framing, not considered here, is to frame AM as an encoder-decoder problem \cite{Bahdanau:2015,Vinyals:2015}. This is an even more general modeling than LSTM-ER. Its suitability for the end-to-end learning task is scope for future work, but its adequacy for component classification and relation identification has been 
investigated in 
\newcite{Potash:2017}. 



\section*{Acknowledgments}
We thank Lucie Flekova, Judith Eckle-Kohler, Nils Reimers, and Christian Stab for valuable feedback and discussions. We also thank the anonymous reviewers for their suggestions. The second author was supported by the German Federal Ministry of Education and Research (BMBF) under the promotional reference 01UG1416B (CEDIFOR).

\bibliography{acl_arg-min-frame2017}
\bibliographystyle{templates/acl_natbib}

\section*{Supplementary Material}
\textbf{Pre-trained word embeddings}: The sequence tagging systems,
including the multi-task learners, as well as the neural dependency parsers can
be initialized with pre-trained word embeddings. For our experiments,
we chose Glove embeddings \cite{Pennington:2014} of different sizes
(50, 100, and 200), the syntactic embeddings of \newcite{Komninos:2016}, and 
the ``structured skip $n$-gram'' model of \newcite{Ling:2015}.
\begin{table*}[t]
\centering
\small
\begin{tabular}{l|c|cc||cc}
  \toprule
  & & \multicolumn{2}{c||}{Paragraph} & \multicolumn{2}{c}{Essay} \\
  & ILP     & LSTM-ER & \texttt{STag}$_{\text{BLCC}}$ & LSTM-ER & \texttt{STag}$_{\text{BLCC}}$\\
  \toprule
  B-C & 51.89 & 59.09 & 50.00 & 56.54 & 53.35 \\
  I-C & 57.74 & 76.09 & 72.46 & 69.67 & 72.72\\ 
  B-MC & 76.56 & 80.64 & 78.26 & 77.15 & 73.80 \\
  I-MC & 55.76 & 58.59 & 50.11 & 59.84 & 54.37 \\ 
  B-P & 62.77 & 77.48 & 74.62 & 73.40 & 75.31\\ 
  I-P & 88.60 & 88.24 & 87.14 & 86.20 & 83.63\\
  O   & 85.74      & 89.08 & 89.52 & 86.65 & 88.81\\ \midrule
  F1 &  68.56      & 75.62 & 71.76 & 72.93 & 72.66\\
  \bottomrule
\end{tabular}
\caption{F1 scores in \% for component segmentation$+$classification. Last row is macro-F1 score.}
\label{table:BIO_TYPE}
\end{table*}

\textbf{Hyperparameter optimization}: Hyperparameter optimization is
an art in itself and often makes the difference between
state-of-the-art results or subpar performance \cite{Wang:2015}. Finding good
parametrizations for neural networks---such as size of the hidden units
or number of hidden layers---is often a very challenging problem.
For the dependency parsers as well as for the sequence taggers $T$ in the \texttt{STag$_{T}$} framing, we performed random hyperparameter optimization \cite{Bergstra:2012}, running 
systems 20 times with hyperparameters randomly chosen within
pre-defined ranges, and then averaged this ensemble of 20 systems.
These ranges were:\footnote{In all cases for the neural networks, we chose a development set of roughly 10\% of the training set.}
\begin{itemize}
        \item BiLSTM tagger in MTL setup: hidden layers of size 150 and 50 dimensional embedding layers (always using 50-dimensional Glove embeddings); the system was trained for 15 iterations and the best model on development set was chosen. All other hyperparameters at their defaults.
        \item BiLSTM-CNN-CRF tagger: one hidden layer of size in $\{125,150,200,250\}$, randomly drawn; training was stopped when performance on development set did not improve for 5 iterations. All other hyperparameters at their defaults. Embeddings randomly chosen from the above-named pre-trained word embeddings, with a preference for 50 dimensional Glove embeddings.
\end{itemize}
For LSTM-ER, we ran the system with 50-dimensional Glove embeddings, which yielded better results than other embeddings we tried, and no further tuning. This is because, as outlined, the system already performs regularization techniques such as entity pre-training and scheduled sampling, which we did not implement for any of the other models. In addition, the system took considerably longer for training, which made it less suitable for ensembling.  

For the neural parsers, our chosen hyperparameters can be read off from the accompanying scripts on our github. We trained the non-neural parsers with default hyperparameters.

\textbf{Practical issues} As outlined in the data section, our
data has a particular structure, but the models we investigate are not
guaranteed to yield outputs that agree with these conditions (unlike,
e.g., ILP models where such constraints can be enforced). For example,
the taggers $T$ in the \texttt{STag$_T$} framing do not need to produce a tree structure, nor do they need to 
produce legitimate B, I, O labeling---e.g., in BIO labeling, an ``I''
may never follow an ``O''. Likewise, while the parsers are guaranteed
to output trees, the labeling they produce need not be consistent with
our data. For example, an argumentative token may be predicted to link
to a non-argumentative unit.
Throughout, we observe very few such violations---that is, the
systems tend to produce 
output consistent with the structures on which they were trained. 
Still, for such violations, we implemented simple and
innocuous 
post-processing rules. 

For the \texttt{STag$_T$} systems, we corrected the following:
\begin{itemize}
  \item[(1)] Invalid BIO structure, i.e., ``I'' follows ``O''.
  \item[(2)] A predicted component is not homogeneous: for example, one token is predicted to link to the following argument component, while another token within the same component is predicted to link to the preceding argument component.
  \item[(3)] A link goes `beyond' the actual text, e.g., when a premise is predicted to link to another component at `too large' distance $|d|$.
\end{itemize}
In case (1), we corrected ``I'' to ''B''. In case (2), we chose the majority labeling within the predicted component. In case (3), we link the component to the maximum permissible component; e.g., when a premise links to a claim at distance 3, but the last component in the document has distance 2, we link the premise to this claim. We applied (1), (2), and (3) in order. For \texttt{STag$_{\text{BLCC}}$} this correction scheme led to 61 out of 29537 tokens changing their labeling in the test data (0.20\%) on essay level and 69 on the paragraph level. For \texttt{STag$_{\text{BL}}$} there were on average many more corrections. For example, 1373 (4.64\%) tokens changed their labeling in the $\mathcal{Y}\text{-3:}\mathcal{Y}_C\text{-3}$ setting described in Table 2. This is understandable because a standard BiLSTM tagger makes output predictions independently; thus, more BIO, etc., violations can be expected.

For the parsers, we additionally corrected when (4) they linked to a non-argumentative unit at index $i_n$. In this case, we would re-direct the faulty link to the ``closest'' component in the vicinity of $i_n$ (measured in absolute distance). Again, we applied (1) to (4) in order. For the LSTM-Parser, this led to 1224 corrections on token level (4.14\%). While this may seem as leading to considerable improvements, this was actually not the case; most of our `corrections' did not improve the measures reported---e.g., token level accuracy decreased, from 57.17\% to 55.68\%. This indicates that a better strategy might have been to re-name the non-argumentative unit to an argumentative unit. 

For LSTM-ER, when a source component is predicted to relate to several targets (something which is always incorrect for our data), we connect the source to its closest target (and no other targets), measured in absolute distance. This is in agreement with the distributional properties of $d$ sketched in Figure 2, which prefers shorter distances over longer ones. 
\subsection*{Links to code used} We used the following code for our experiments: BLCC ({https://github.com/XuezheMax/LasagneNLP}); MTL BL ({https://bitbucket.org/soegaard/mtl-cnn/src}); LSTM-ER ({https://github.com/tticoin/LSTM-ER}); LSTM-Parser ({https://github.com/clab/lstm-parser}); Kiperwasser parser ({https://github.com/elikip/bist-parser}); Mate parser ({https://code.google.com/archive/p/mate-tools/wikis/ParserAndModels.wiki}); MST parser ({http://www.seas.upenn.edu/ ~strctlrn/MSTParser/MSTParser.html}). The results for the ILP model were provided to us by the first author of \newcite{Stab:2016}.

\section*{Error Analysis}
We conduct some more error analysis, focussing on the three best models ILP, LSTM-ER and \texttt{STag$_{\text{BLCC}}$}. 

Which component \emph{types} are particularly difficult to detect? 
Table \ref{table:BIO_TYPE} investigates F1-scores for component segmentation$+${classification}. In this case, there are seven classes: $\{\text{B},\text{I}\}\times\{\text{C},\text{MC},\text{P}\}\cup\{\text{O}\}$. We observe that the O class is particularly easy, as well as I-P. These two are the most frequent labels in the data and are thus most robustly estimated. 
While all systems are more troubled predicting the beginning of a claim than its continuation (this is often due to difficulty of predicting the inclusion or omission of discourse markers as illustrated above), major claims follow a reverse trend. Further analysis reveals that claims are often mistaken for premises and vice versa, and major claims for claims or---to a lesser degree---for premises.
The mismatch between claims and premises is sometimes due to misleading introductory phrases such as ``\emph{Consequently ,}'' which often imply conclusions (and hence claims), but sometimes also give reasons---i.e., premises---for other claims or premises.

We also note that the ILP model is substantially worse than the two LSTMs in all cases except for I-P on the component segmentation$+$classification task. 

A major source of errors for \emph{relations} is that either of their arguments (the two components) do not match exactly or approximately. When they do match, errors are mostly a mismatch between actual Attack/Against vs.\ predicted Support/For relations. Support/For relations are the vast majority in the PE data (94\% and 82\%, respectively). In rare cases, the two arguments have been correctly identified but their types are wrong (e.g.\ premise and claim while the gold components are claim and major claim, respectively).

\end{document}